\documentclass[journal]{IEEEtran}
\IEEEoverridecommandlockouts

\usepackage{pgfplots}
\pgfplotsset{compat=newest}

\usepackage{enumitem}
\usepackage{array}
\usepackage{multirow}
\usepackage{rotating}
\usepackage{caption}
\usepackage{subcaption}
\usepackage{adjustbox}
\usepackage{cite}
\usepackage{hyperref}
\usepackage{amsmath,amssymb,amsfonts}
\usepackage{algorithmic}
\usepackage{graphicx}
\usepackage{textcomp}
\usepackage{xcolor}
\usepackage{array}
\def\BibTeX{{\rm B\kern-.05em{\sc i\kern-.025em b}\kern-.08em
    T\kern-.1667em\lower.7ex\hbox{E}\kern-.125emX}}
    
\newcommand{\printfnsymbol}[1]{%
  \textsuperscript{*}%
}    

\newcommand{\ie}{\textit{i}.\textit{e}., }
\newcommand{\eg}{\textit{e}.\textit{g}., }

\begin{document}

\title{ECG Biometric Recognition: Review, System Proposal, and Benchmark Evaluation}

\author{Pietro Melzi, Ruben Tolosana, Ruben Vera-Rodriguez \\
Biometrics and Data Pattern Analytics - BiDA Lab, Universidad Autonoma de Madrid \\
{\tt\small \{pietro.melzi, ruben.tolosana, ruben.vera\}@uam.es}
}

\maketitle

\begin{abstract}
Electrocardiograms (ECGs) have shown unique patterns to distinguish between different subjects and present important advantages compared to other biometric traits, such as difficulty to counterfeit, liveness detection, and ubiquity. Also, with the success of Deep Learning technologies, ECG biometric recognition has received increasing interest in recent years. However, it is not easy to evaluate the improvements of novel ECG proposed methods, mainly due to the lack of public data and standard experimental protocols. In this study, we perform extensive analysis and comparison of different scenarios in ECG biometric recognition. Both verification and identification tasks are investigated, as well as single- and multi-session scenarios. Finally, we also perform single- and multi-lead ECG experiments, considering traditional scenarios using electrodes in the chest and limbs and current user-friendly wearable devices.

In addition, we present ECGXtractor, a robust Deep Learning technology trained with an in-house large-scale database and able to operate successfully across various scenarios and multiple databases. We introduce our proposed feature extractor, trained with multiple sinus-rhythm heartbeats belonging to 55,967 subjects, and provide a general public benchmark evaluation with detailed experimental protocol. We evaluate the system performance over four different databases: \textit{i)} our in-house database, \textit{ii)} PTB, \textit{iii)} ECG-ID, and \textit{iv)} CYBHi. With the widely used PTB database, we achieve Equal Error Rates of 0.14\% and 2.06\% in verification, and accuracies of 100\% and 96.46\% in identification, respectively in single- and multi-session analysis. We release the source code, experimental protocol details, and pre-trained models in GitHub to advance in the field.
\end{abstract}


\section{Introduction}
Biometric data are widely used in recognition systems due to their ability to uniquely identify subjects through their biological or behavioural traits~\cite{jain_50_2016}, which are intrinsic to human beings, differently from traditional recognition tools such as passwords and tokens. Among the most popular biometric traits, we encounter facial features~\cite{morales_sensitivenets_2020}, fingerprints~\cite{priesnitz_mobile_2022}, iris~\cite{mostofa_deep_2021}, gait~\cite{delgado-santos_gaitprivacyon_2021}, handwriting~\cite{tolosana_biotouchpass2_2020}, and speech~\cite{alharbi_automatic_2021}. However, these traditional biometric traits are vulnerable against Presentation Attacks (PAs)~\cite{george_learning_2021, czajka_presentation_2018, tolosana_biometric_2019, tolosana_presentation_2019} and digital manipulations~\cite{2021_Book_DigitalFaceManipulation}.

The electrocardiogram (ECG) is a graph that reproduces the electrical activity of the heart, obtained by placing electrodes over suitable parts of the body. Even if its deployment in real applications is not as popular as most established biometric traits, ECG presents interesting advantages for biometric recognition. First of all, ECGs provide higher security as the signal is measured inside the body, which is therefore difficult to simulate or copy~\cite{karimian_vulnerability_2017}. ECGs allow liveness detection, as they can be captured from living subjects only, and provide useful additional information related to psychological states and clinical status~\cite{melzi_analyzing_2021}. The possibility to capture ECGs from fingers~\cite{islam_biometric_2017} or through wearable devices~\cite{steinberg_novel_2019} simplifies their acquisition and increases the acceptability of ECG signals for commercial and public applications~\cite{alduwaile_using_2021}.

However, biometric systems based on ECG have not yet reached the same level of technological maturity and acceptance compared to other biometric traits, mainly because of the lack of public databases~\cite{ingale_ecg_2020}. In addition, it is not easy to evaluate the improvements of novel proposed approaches, as different databases and experimental protocols are usually considered. Also, multiple ECG signals collected over time from the same subject present a certain variability between them, \eg due to mental, emotional, or physical changes, or permanently due to changes in lifestyle or individual characteristics~\cite{odinaka_ecg_2012}. Extensive studies recognize that heart rate variability may be induced by heart diseases, mental and physical variations, and other factors including drugs, medications, and diet~\cite{rajendra_acharya_heart_2006}. Hence, experimental protocols that consider for each subject ECG signals recorded during different time sessions are necessary to assess effectiveness and robustness of ECG-based recognition systems. However, a large number of previous studies performs experiments based on single-session analysis~\cite{alduwaile_using_2021, li_toward_2020, donida_labati_deep-ecg_2019}, which are not very informative of the system performance in realistic scenarios. In this study we address the issue of ECG variability over time, investigated through multi-session experiments where we specify the distance in time between ECGs belonging to the same subjects and recorded in different sessions.

In this study we consider two different biometric recognition scenarios based on ECG signals: \textit{i)} verification and \textit{ii)} identification. We develop a comprehensive technology, ECGXtractor, based on Deep Learning (DL) systems and trained with a large-scale in-house database, to investigate both scenarios. In particular, we make use of an Autoencoder to extract discriminative features from ECG signals. The Autoencoder is successfully applied to a variety of ECG signals featured with different properties. It is important to note that our Autoencoder is not specifically trained to extract features suitable for biometric recognition tasks. Hence, the extracted features may also be used in additional applications, \eg the prediction of health conditions.

The main contributions of this study are:
\begin{itemize}
\item In-depth analysis of state-of-the-art DL approaches for ECG biometric recognition, highlighting the key aspects of the scenarios considered in current real applications. 

\item Proposal of ECGXtractor, a new DL-based system suitable for different tasks based on ECG time signals, composed of a general feature extractor that can be used for different recognition tasks, \ie identification and verification. The feature extractor consists in an Autoencoder trained with a large set of 166,932 heartbeats collected in an in-house database and extracted from the ECG signals belonging to 55,967 subjects.

\item Extensive analysis and comparison of different biometric recognition scenarios based on ECGs. To evaluate the overall impact of different conditions and experimental settings, both verification and identification tasks are investigated, as well as single- and multi-session acquisition scenarios. We evaluate our proposed ECGXtractor with different ECG databases, that may contain single- or multi-lead ECG signals, recorded with ``on-the-person'' or ``off-the-person'' technologies, \ie sensors that require conductive paste or gel when attached to body surfaces, or do not require any special preparation of the subject with objects or surfaces~\cite{ingale_ecg_2020}. We present to the research community a clear and structured experimental protocol and benchmark, that allows to easily reproduce our experiments and overcome some drawbacks existing in other studies. 

\item We release the source code, experimental protocol details, and the trained weights of the ECGXtractor approach in GitHub\footnote{\url{https://github.com/BiDAlab/ECGXtractor}} to advance in the field.

\end{itemize}

To the best of our knowledge, this study simultaneously addresses for the first time a variety of key aspects, which includes: \textit{i)} the specific recognition task, \textit{ii)} the variability of ECG signals over time, \textit{iii)} the number of leads available, and \textit{iv)} the modality of recording.

The remainder of the paper is organised as follows. Sec.~\ref{sec:related_works} summarizes previous studies carried out in the field of ECG biometric recognition. Sec.~\ref{sec:proposed_method} explains the details of our proposed approach: ECGXtractor, which set the foundations for our experiments. Sec.~\ref{sec:databases} describes the main details of the databases considered in the study. Sec.~\ref{sec:verification} and Sec.~\ref{sec:identification} respectively refer to the tasks of verification and identification, including both experimental protocols and results. Finally, Sec.~\ref{sec:conclusion} draws the conclusions of this study and points out some lines for future work.

\section{Related Works}\label{sec:related_works}
In the literature, many studies have been conducted on ECG biometric recognition, considering different experimental settings depending on the task (\ie verification or identification), the number of ECG recordings per user and leads available, the type of recorder (\ie ``on-the-person'' or ``off-the-person''), and other additional aspects related to data pre-processing and feature extraction. Because of that, standard experimental protocols are missing and difficulties arise when comparing novel approaches with the state of the art. 

In this sense, Ingale \textit{et al.} evaluated with multiple databases the effectiveness of various techniques applied in different phases of ECG biometric recognition, \ie feature extraction, signal filtering, segmentation, and matching. This study was motivated by the fact that most of the methods proposed in the literature fails to report standard metrics~\cite{ingale_ecg_2020}. 

With regard to the properties of ECG signals, in the literature a variety of lead configurations and modalities of recording have been considered. According to~\cite{odinaka_ecg_2012}, a single lead contains sufficient information to support biometric recognition. However, some studies adopted multiple leads in an effort to improve performance~\cite{krasteva_biometric_2017, krasteva_perspectives_2018}. Furthermore, an increasing number of portable devices, such as fingertips or wearable devices with dry electrodes, allows to record ``off-the-person'' ECG signals in a non-intrusive way. ``Off-the-person'' recording provides more noise and variability compared to traditional ``on-the-person'' recording~\cite{ingale_ecg_2020}, but it is considered more reasonable and aligned with industrial requirements when using ECG signals for biometric recognition~\cite{bansal_portable_2018, haverkamp_accuracy_2019}.

In this section, we analyze state-of-the-art ECG biometric systems based on DL technologies. In general, DL systems extract features from ECG signals in a convenient and reliable manner, and provide better results compared to traditional handcrafted systems, \eg based on Support Vector Machines (SVMs)~\cite{liu_multiscale_2018, paiva_beat-id_2017, komeili_feature_2018}. In Table~\ref{table:related_works} we summarize the studies discussed in this section, reporting their main characteristics and performances. We observe that PTB~\cite{bousseljot_nutzung_2009} and ECG-ID~\cite{lugovaya2005} are the most popular databases in the literature and, for the former, several studies only focus on its subset of healthy subjects. In addition, multi-session acquisition analysis is not always addressed in the literature: some studies fail to report any information about it, or consider subjects provided with single ECG signals.

\begin{table*}[t]
\centering
\caption{Comparison of different deep learning approaches for ECG biometric recognition. Some studies do not specify if their experiments are conducted according to single- or multi-session acquisition analysis. In such cases, we expect that multi-session is not considered. It is also possible that for some subjects only a single ECG signal is available in the database. Acc = accuracy, CWT = Continuous Wavelet Transform, EER = Equal Error Rate.}
\scalebox{1.00}{\renewcommand{\arraystretch}{1.2}%
\begin{tabular}{cccccc}
\textbf{Study}                      & \textbf{Input}                           & \textbf{Task}                                                                                                                               & \textbf{Database}                            & \textbf{Performance}   & \textbf{Session}                 \\ \hline\hline
\multirow{3}{*}{\begin{tabular}[c]{@{}c@{}} AlDuwaile \textit{et al.} (2021)\\~\cite{alduwaile_using_2021} \end{tabular}} & \multirow{3}{*}{CWT image from heartbeat}                                                                                                & \multirow{3}{*}{Identification} & PTB                                & Acc = 100\%   & Not specified                  \\ \cline{4-6} 
                           &                                                                                                                                          &                                 & \multirow{2}{*}{ECG-ID}            & Acc = 99.33\% & Single                         \\ \cline{5-6} 
                           &                                                                                                                                          &                                 &                                    & Acc = 97.28\% & Multi                          \\ \hline
\multirow{4}{*}{\begin{tabular}[c]{@{}c@{}} Chu \textit{et al.} (2019)\\~\cite{chu_ecg_2019} \end{tabular}}       & \multirow{4}{*}{Two heartbeats}                                                                                                          & \multirow{2}{*}{Identification} & ECG-ID                             & Acc = 98.24\% & \multirow{4}{*}{Not specified} \\ \cline{4-5}
                           &                                                                                                                                          &                                 & PTB (healthy subjects)             & Acc = 100\%   &                                \\ \cline{3-5}
                           &                                                                                                                                          & \multirow{2}{*}{Verification}   & ECG-ID                             & \textbf{EER = 2.00\%}  &                                \\ \cline{4-5}
                           &                                                                                                                                          &                                 & PTB (healthy subjects)             & EER = 0.59\%  &                                \\ \hline
\multirow{3}{*}{\begin{tabular}[c]{@{}c@{}} Donida \textit{et al.} (2019)\\~\cite{donida_labati_deep-ecg_2019} \end{tabular}}    & \multirow{3}{*}{Vector of QRS complexes}                                                                                                 & Identification                  & PTB (healthy subjects)             & Acc = 100\%   & Not specified                \\ \cline{3-6} & & \multirow{2}{*}{Verification}   & \multirow{2}{*}{E-HOL-03-0202-003} & EER = 1.05\%  & Single (within 300 s)           \\ \cline{5-6} 
                           &                                                                                                                                          &                                 &                                    & EER = 2.15\%  & Single (within 500 min)        \\ \hline
\multirow{2}{*}{\begin{tabular}[c]{@{}c@{}} Ihsanto \textit{et al.} (2020)\\~\cite{ihsanto_fast_2020} \end{tabular}}   & \multirow{2}{*}{Six to eight heartbeats}                                                                                                 & \multirow{2}{*}{Identification} & ECG-ID                             & \textbf{Acc = 100\%}   & Multi (same day)               \\ \cline{4-6} 
                           &                                                                                                                                          &                                 & MIT-BIH                            & Acc = 100\%   & Single                         \\ \hline
\multirow{3}{*}{\begin{tabular}[c]{@{}c@{}} Ingale \textit{et al.} (2020)\\~\cite{ingale_ecg_2020} \end{tabular}}    & \multirow{3}{*}{\begin{tabular}[c]{@{}c@{}}Template achieved with \\ both fiducial and non-fiducial \\ feature extraction\end{tabular}} & \multirow{3}{*}{Verification}   & ECG-ID                             & \textbf{EER = 2.00\%}  & \multirow{3}{*}{Not specified} \\ \cline{4-5}
                           &                                                                                                                                          &                                 & PTB                                & \textbf{EER = 0.25\%}  &                                \\ \cline{4-5}
                           &                                                                                                                                          &                                 & CYBHi (126 subjects)               & \textbf{EER = 2.30\%}  &                                \\ \hline
\multirow{3}{*}{\begin{tabular}[c]{@{}c@{}} Li \textit{et al.} (2020)\\~\cite{li_toward_2020} \end{tabular}} & \multirow{3}{*}{\begin{tabular}[c]{@{}c@{}} Multivote of\\single heartbeats \end{tabular}}  & \multirow{3}{*}{Identification}                  & FANTASIA                           & Acc = 100\%   & Single                         \\ \cline{4-6}
 & & & \begin{tabular}[c]{@{}c@{}} FANTASIA, CEBSDB,\\NSRDB, STDB, AFDB \end{tabular} & \begin{tabular}[c]{@{}c@{}}Acc = 97.7\% \\(Average) \end{tabular} & Not specified \\ \hline
\multirow{2}{*}{\begin{tabular}[c]{@{}c@{}} Srivastva \textit{et al.} (2021)\\~\cite{srivastva_plexnet_2021}\end{tabular}} & \multirow{2}{*}{Three heartbeats}                                                                                                        & \multirow{2}{*}{Identification} & PTB                                & \textbf{Acc = 99.66\%} & Multi                          \\ \cline{4-6} 
                           &                                                                                                                                          &                                 & CYBHi (63 subjects)                & \textbf{Acc = 99.66\%} & Multi                          \\ \hline
\multirow{5}{*}{\textbf{Proposed}} & \multirow{5}{*}{\begin{tabular}[c]{@{}c@{}} \textbf{Template achieved}\\ \textbf{with multiple}\\ \textbf{heartbeats} \end{tabular}} & \multirow{3}{*}{\textbf{Verification}} & \textbf{PTB (healthy subjects)} & \textbf{EER = 0.00\%} & \textbf{Single} \\ \cline{4-6}
& & & \textbf{PTB} & \textbf{EER = 2.06\%} & \textbf{Multi} \\ \cline{4-6}
& & & \textbf{ECG-ID} & \textbf{EER = 0.15\%} & \textbf{Multi (same day)} \\ \cline{3-6}
& &\multirow{2}{*}{\textbf{Identification}} & \multirow{2}{*}{\textbf{PTB}} & \textbf{Acc = 100\%} & \textbf{Single} \\ \cline{5-6}
& & & & \textbf{Acc = 96.46\%} &\textbf{Multi} \\ \hline
 
\end{tabular}
}
\label{table:related_works}
\end{table*}

The first DL system that we discuss is the Cascaded Convolutional Neural Network (CNN), proposed in~\cite{li_toward_2020}. This system is composed of two CNNs: \textit{i)} F-CNN, for feature extraction, and \textit{ii)} M-CNN, for feature comparison. Single heartbeats extracted from single-lead ECG signals were given as input to the F-CNN. Match/non-math binary outcomes were provided by M-CNN for the comparison of features extracted from paired ECG signals. For each heartbeat in evaluation, the subject whose template provided the highest matching scores was the predicted identity. The Cascaded CNN was evaluated on five databases and achieved up to 100\% of accuracy when trained and tested with the FANTASIA database~\cite{iyengar_age-related_1996}, which contains single-session ECG signals from 40 healthy subjects. This architecture is more scalable than other identification systems, as it does not require to re-train the model when new subjects are considered. 

In~\cite{ihsanto_fast_2020}, a Residual CNN was separately trained twice for identification with two different sets of 90 and 48 subjects. Single-lead databases and single-session (or multi-session but recorded during the same day) ECG signals were employed. Accuracy of 100\% was achieved when considering multiple heartbeats for each subject. The limitations of studies that only consider single-session databases were pointed out in~\cite{alduwaile_using_2021}. In that study, single segments were extracted from single-lead ECG signals and provided as input to different CNNs. The accuracy decreased from single- to multi-session analysis: accuracies of 100\% and 99.33\% were achievable with two different databases in single-session, while 97.28\% was the highest accuracy achievable in multi-session. Finally, an ensemble of state-of-the-art pre-trained deep neural networks for identification was proposed in~\cite{srivastva_plexnet_2021}. Segments of three consecutive heartbeats were extracted from ECG signals and provided as input to the system. By taking advantages of both transfer learning and ensemble learning, such system achieved an accuracy of 99.66\%, also considering multi-session recordings.

Regarding verification, we have already described the work of Ingale \textit{et al.}~\cite{ingale_ecg_2020}. In that study, some databases provided single ECG signals for their subjects and multi-session acquisition was not taken into account. In~\cite{donida_labati_deep-ecg_2019}, a system composed of a CNN extracting features from multiple single-lead QRS complexes combined together was proposed. Equal Error Rates (EERs) of 1.05\% and 2.26\% were achieved when considering QRS complexes extracted respectively within 300 s and 150 min from a 24 hour-long recording. Finally, we consider the parallel multi-scale one-dimensional residual network proposed in~\cite{chu_ecg_2019}, trained with a special loss function to extract features that improve the generalization ability and achieve more stable results across databases. In that study, the input data of the neural network were heartbeat vectors, composed of two single-lead heartbeats randomly selected from each subject. The authors achieved a 0.59\% EER on the subset of healthy subjects contained in the PTB database. For most of these subjects, only a single ECG recording was provided and, hence, multi-session experiments could not be performed.

As emerged in this section, there are no common experimental protocols to realize and compare state-of-the-art technologies for ECG biometric recognition. The various DL systems proposed in the literature require as input heartbeats segmented and/or combined according to different procedures. Some of these procedures may be particularly suitable for specific databases and do not generalize well with others. Also, DL systems are evaluated on different scenarios featured with proper experimental settings, which makes any comparison difficult. Furthermore, to the best of our knowledge, the most realistic scenario of multi-session biometric verification is not sufficiently investigated in the literature. Existing studies on biometric verification do not specify whether the processed ECG samples are recorded in the same or in different sessions~\cite{chu_ecg_2019, ingale_ecg_2020}, or consider databases providing only a single ECG signal for each subject~\cite{donida_labati_deep-ecg_2019}. 

\begin{figure*}[tb]
\centering
\centerline{\includegraphics[width=1\linewidth]{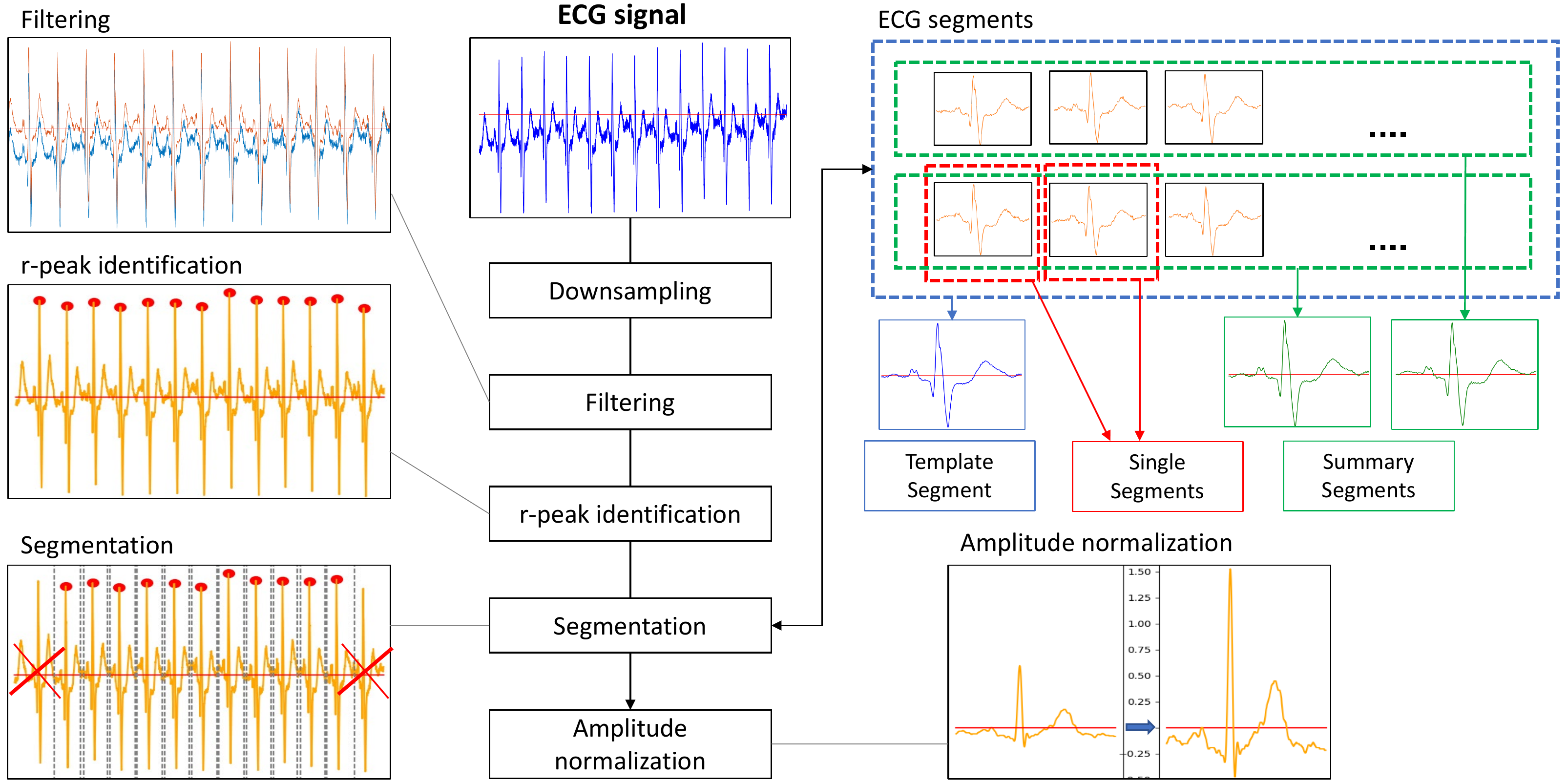}}
\caption{Graphical representation of the pre-processing operations of ECGXtractor performed on ECG signals to obtain template segments, single segments, and summary segments for ECG biometric recognition with normalized amplitude. Template segments are generated from all the single segments identified in the original ECG signal. Summary segments are generated from blocks of ten consecutive single segments (color image).}
\label{fig:proposed_method_1}
\end{figure*}

To overcome all these issues, we provide in this study an in-depth analysis and benchmark of different experimental settings for ECG biometric recognition. We propose ECGXtractor, a DL method able to successfully perform biometric recognition in multiple scenarios, and with multiple databases. We make our system publicly available, so that it will be easy for the research community to reproduce our experiments. Moreover, a significant focus of our study is dedicated to the relevant scenario of multi-session biometric verification.

\begin{figure*}[tb]
\centering
\centerline{\includegraphics[width=1\linewidth]{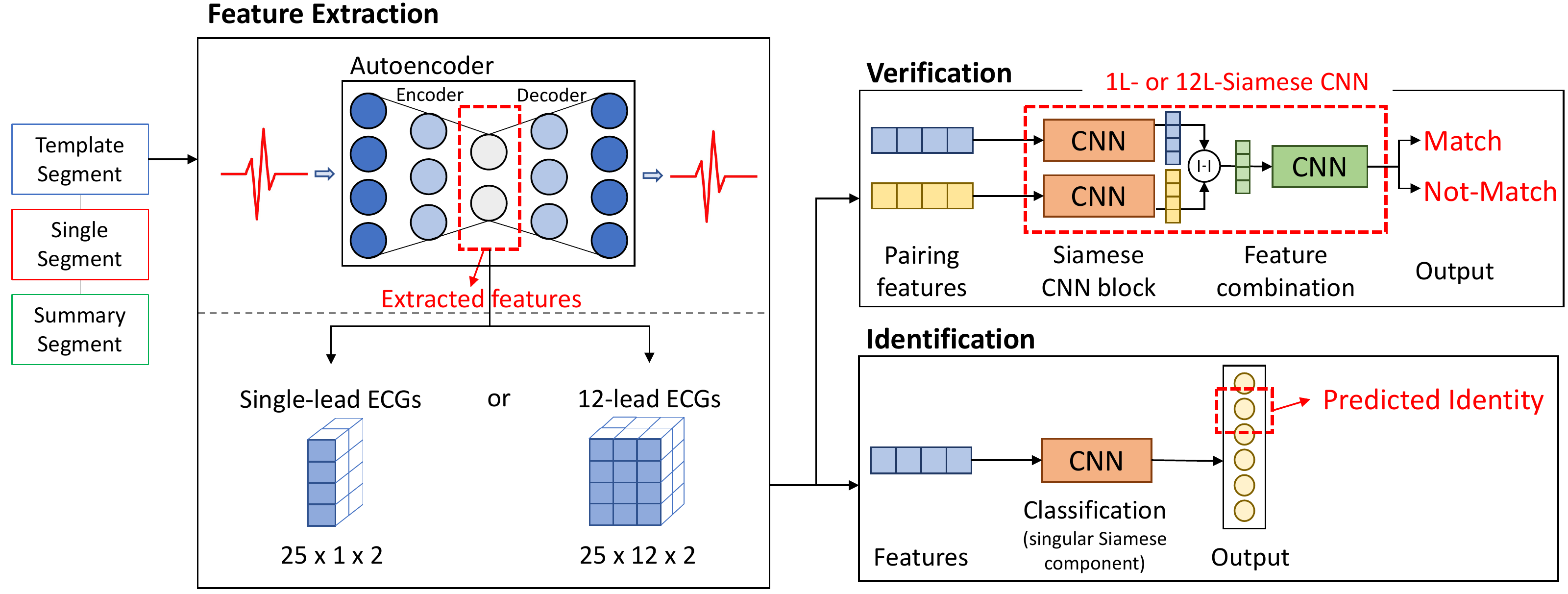}}
\caption{Graphical representation of the feature extraction considered in ECGXtractor, performed on the different types of segments to carry out ECG biometric verification and identification. Features are extracted from the latent feature representation of an Autoencoder and, in case of verification, the creation of pairs of features (\ie genuine-genuine and genuine-impostor pairs) is required before any further operation (color image).}
\label{fig:proposed_method_2}
\end{figure*}

\section{Proposed Method}\label{sec:proposed_method}
Fig.~\ref{fig:proposed_method_1} and Fig.~\ref{fig:proposed_method_2} describe the pre-processing and feature extraction of ECGXtractor, our proposed approach for ECG biometric recognition. The source code and pre-trained models are available in GitHub\footnote[1]{\url{https://github.com/BiDAlab/ECGXtractor}}. Multiple ECG databases have been considered in this study to investigate how different properties of ECG signals may affect performance in ECG biometric recognition. For this reason, a set of preliminary operations is required to mitigate the discrepancies existing between ECG time signals recorded with different sensors. In particular, these preliminary operations allow us to train our DL system with a large-scale in-house database, and exploit the generated knowledge with multiple smaller databases.

\subsection{Pre-processing}
ECG signals from different databases are recorded with frequencies of 1 KHz or 500 Hz. We downsample all of them to 500 Hz. We also apply Finite Impulse Response Filters to our ECG signals, to maintain frequencies between 0.7 and 90 Hz and remove those frequencies around 50 Hz, noisy due to power supply. Subsequently, for each ECG signal we identify its r-peaks through the reliable method \textit{ecg\_peaks}, implemented in the \textit{neurokit2} toolbox~\cite{Makowski2021neurokit}, and discard its first and last r-peaks. Then, we build single segments, \ie single heartbeats centered around each r-peak. According to a previous work~\cite{li_toward_2020}, and considering the average heartbeat length of 0.8 s, we fix the length of our single segments to 0.32 s before and 0.48 s after the r-peak. In case of multi-lead ECG signals, we identify r-peaks on the signal recorded with Lead I and build single segments containing the multi-lead recording of single heartbeats. In our proposed method, the following approaches are studied:
\begin{itemize}
	\item template segments (templates along the paper), \ie ECG segments with the shape of single heartbeats, obtained from the processing of all the single segments contained in an ECG signal.
	\item summary segments, \ie ECG segments with the shape of single heartbeats, obtained from the processing of ten consecutive single segments contained in an ECG signal.
\end{itemize} 
For template generation, we adopt a procedure similar to the one presented in~\cite{li_toward_2020}, here described for a generic ECG signal:

\begin{enumerate}
\item At first, all the segments representing single heartbeats are identified by means of the segmentation described above. These single segments contain 400 time samples for each lead.
\item The element-wise average of the identified single segments is computed, separately for each ECG lead acquired.
\item The five single segments presenting the smallest Euclidean distance from the element-wise average segment are identified.
\item The five identified single segments are element-wise averaged, separately for each lead, to obtain the final template.
\end{enumerate}

This procedure aims to minimize the effect of noise contained in ECG signals. The same steps are applied to generate summary segments, with the only difference that blocks of ten consecutive single segments are considered in step 2) instead of entire ECG signals. As a consequence, multiple summary segments and only one template can be obtained from single ECG signals.

Finally, the amplitude of the ECG segments, regardless of whether they are templates, single segments, or summary segments, is normalized to a fixed value of 2 mV, multiplying every time sample by the ratio between 2 mV and the current amplitude of the segment. The operation, performed separately for each lead, aims to eliminate the amplitude variability existing between the different databases. In Fig.~\ref{fig:proposed_method_1} we provide a summary of the described operations.

\subsection{Feature Extraction}\label{subsec:feature_extractor}
To extract features from ECG segments (\ie template, single segment, and summary segment), we consider a DL-based Autoencoder, \ie a neural network composed of two parts: \textit{i)} an encoder, that reduces the size of the input data and learns its encoded representation, and \textit{ii)} a decoder, that attempts to reconstruct the input data from the encoded representation. In the context of ECGs, Autoencoders can be applied to many tasks, such as noise reduction and heartbeat type classification~\cite{nurmaini_deep_2020} or lower dimensional representation and biometric recognition~\cite{eduardo_ecg-based_2017}. In this study, we make use of the Autoencoder for the latter task.

\begin{figure}[tb]
\centering
\centerline{\includegraphics[width=1\linewidth]{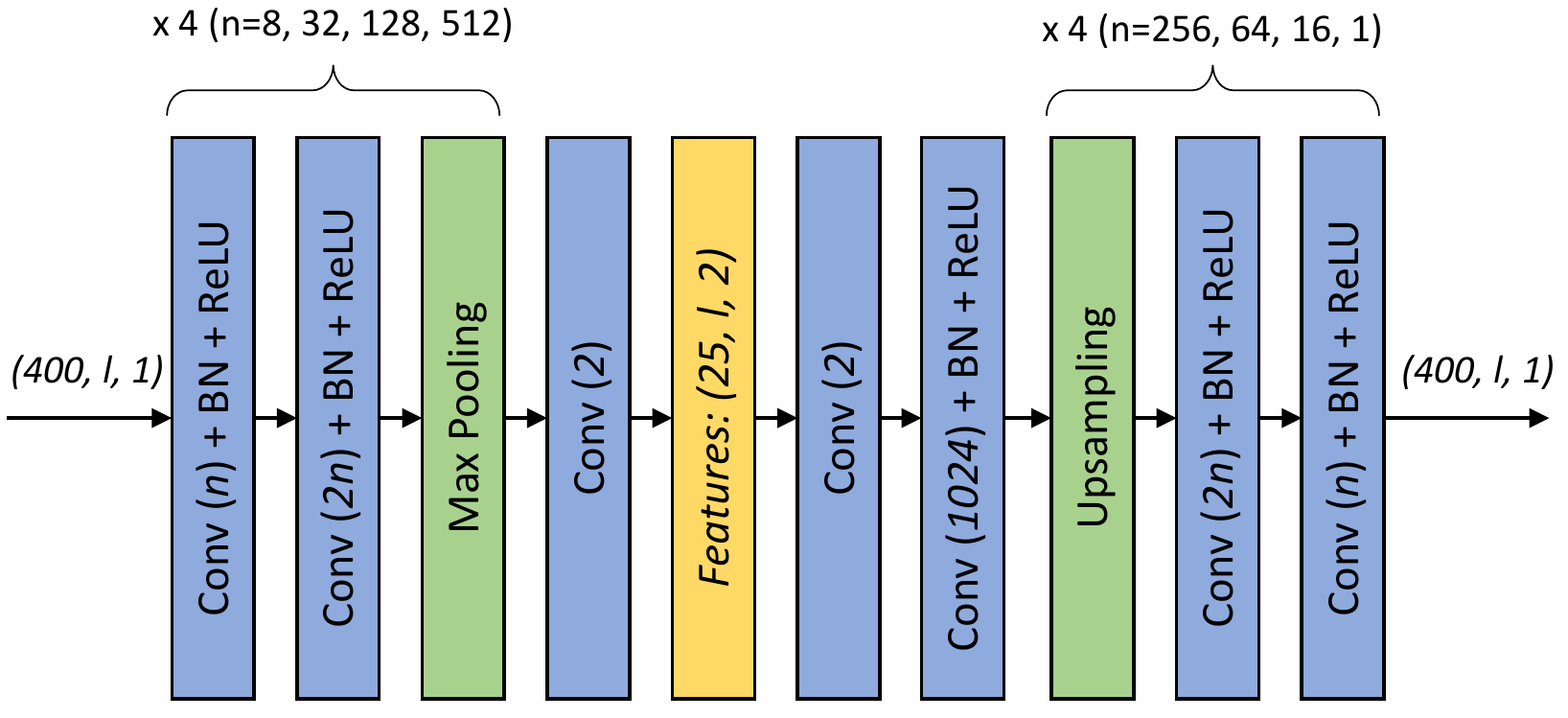}}
\caption{Architecture of Autoencoder. \textit{l} = leads, BN = Batch Normalization, ReLU = Rectified Linear Unit (color image).}
\label{fig:autoencoder}
\end{figure}

Our Autoencoder (Fig.~\ref{fig:autoencoder}) consists in a modified version of the Variational Autoencoder proposed in~\cite{kuznetsov_interpretable_2021} for feature extraction and synthetic heartbeats generation. We only maintain the convolutional layers from the original architecture, applying small changes to them and discarding the fully connected and variational components. We observed that the features extracted with the original architecture were not intended for biometric recognition. Our Autoencoder requires as input a time signal that represents an ECG segment, \ie template, single segment, or summary segment. Our encoder is composed of four groups of convolutional, batch normalization, ReLU activation, and max pooling layers. After the last max pooling layer, a further convolutional layer with two output channels is added, to reduce the size of the latent features that will be extracted. Our decoder presents an almost symmetric architecture, to reconstruct the ECG segment provided as input to the encoder. It is important to point out that features are extracted independently from each lead, \ie numerical values from different leads are never combined to generate features. Hence, the same pre-trained Autoencoder is applied for feature extraction for both 1-Lead and 12-Lead ECG segments, which is a considerable advantage for real applications.

We exploit the Autoencoder to extract features from each segment of interest, \ie templates, single segments, and summary segments. According to the architecture of our Autoencoder, two channels of 25 temporal features each are extracted from each lead of the segment provided as input. Experimental trials showed the advantages of considering two channels instead of one. Hence, the size of the features generated from each segment is $25 \times l \times 2$, where \textit{l} is the number of leads considered. In our experiments, we consider single-lead segments ($l=1$) and 12-lead segments ($l=12$). We observe that, by building templates or summary segments and extracting features from them, we achieve a considerable data minimization compared to the size of original ECG signals, without losing data usefulness. The extracted features are considered in both recognition tasks investigated in the study: verification and identification. In Fig.~\ref{fig:proposed_method_2}, the key aspects of both ECG biometric recognition tasks are presented.

\subsection{ECG Biometric Verification}\label{subsec:ecgbiomver}

\begin{figure}[tb]
\centering
\centerline{\includegraphics[width=1\linewidth]{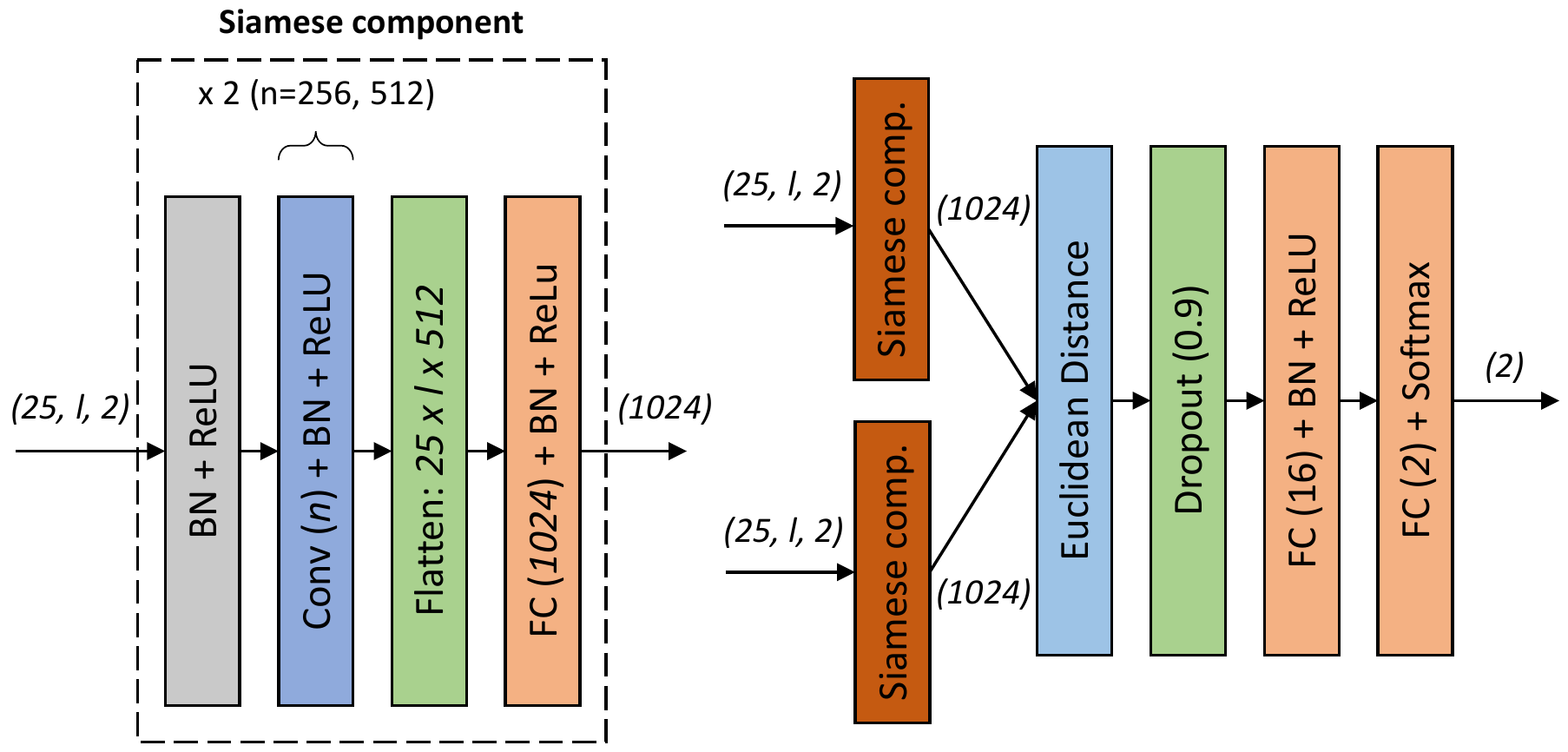}}
\caption{Architecture of 1L- and 12L-Siamese CNN. \textit{l} = leads, BN = Batch Normalization, FC = Fully Connected, ReLU = Rectified Linear Unit (color image).}
\label{fig:siamese}
\end{figure}

In ECG biometric verification, features extracted from different leads are combined together during their processing. For this reason, two Siamese Convolutional Neural Networks (Siamese CNNs) are trained for the task of verification, one for single-lead ECG signals (1L-Siamese CNN) and one for 12-lead ECG signals (12L-Siamese CNN). In case of single-lead ECGs, we consider signals recorded with Lead I, as this is the typical lead available in ECG databases and the most basic lead measured by smartwatches and other wearable devices~\cite{han_automated_2021}. The two Siamese CNNs require as input pairs of feature vectors extracted from two different ECG segments, and predict if such segments belong to the same subject or not. 1L- and 12L-Siamese CNNs are depicted in Fig.~\ref{fig:siamese}. For each pair of feature vectors, the Siamese component generates two vectors of 1024 features. The Euclidean distance is calculated for each of the 1024 features, and the resulting vector is processed to provide final match/not-match decisions.

The two Siamese CNNs are trained and evaluated with genuine and impostor pairs of features extracted from different ECG segments. For each subject considered during training, we build their template with the first available ECG and extract single segments from their remaining ECGs. By matching templates with single segments of same and different subjects, we generate genuine and impostor pairs for training. We consider single segments less reliable than templates and summary segments to represent the intra-user variability. For this reason we employ single segments only during training, to provide our system with the knowledge derived from potentially more challenging pairs, and not during evaluation, where summary segments and templates are considered respectively in single- and multi-session scenarios.

We evaluate 1L- and 12L-Siamese CNNs in both single- and multi-session acquisition scenarios. In single-session scenario we have the constraint to dispose of only one ECG signal for each subject. Hence, we divide the considered ECG signal in blocks of ten consecutive single segments, and generate a summary segment from each block. By randomly matching summary segments of same and different subjects, we create genuine and impostor pairs for evaluation. We consider multi-session verification the most important scenario of this study, as it represents the most realistic situation for commercial and widespread biometric recognition technologies. In multi-session verification the system performance may be negatively affected by intra-user variability, occurring when biometric traits of the subject change over time~\cite{tolosana_reducing_nodate}. To investigate this aspect, we generate enrolment templates with the first ECG signal of each subject, and probe templates with the other available ECGs acquired in other time sessions, to create realistic genuine/impostor pairs for evaluation.

\subsection{ECG Biometric Identification} 

To assess the validity of the features extracted with our Autoencoder and compare our DL system with others studies in the literature, we also design experiments in the scenarios of single- and multi-session identification. We consider the same trained singular Siamese component of our 1L- or 12L-Siamese CNN for the verification task as described in Sec.~\ref{subsec:ecgbiomver}, according to the number of leads of input data, and include at the end a new fully connected layer with output dimension that varies according to the number of subjects considered in each experiment. This is the only layer to train for the identification task. For all the experiments of identification we consider summary segments generated from blocks of ten consecutive single segments, so that we have enough samples to train and evaluate our system.

\section{Databases}\label{sec:databases}
Four different databases have been considered in this study. They contain single- and multi-lead ECG signals, recorded with ``on-the-person'' or ``off-the-person'' modality. With these four databases, we evaluate the impact of different ECG properties on biometric recognition. We use an in-house database~\cite{sanz-garcia_electrocardiographic_2021, melzi_analyzing_2021}, and three public databases widely used in the literature, to compare the performances achieved in our experiments with other studies. The main characteristics of the databases are reported in Table~\ref{table:databases}.
\newline

\begin{table}[tb]
\centering
\caption{Summary of the main characteristics of the considered databases. On-TP = ``on-the-person'', Off-TP = ``off-the-person'', SR = Sinus Rhythm.}
\begin{adjustbox}{width=0.455\textwidth}\renewcommand{\arraystretch}{1.2}%
\begin{tabular}{cccccc}
\textbf{Database} & \textbf{Leads} & \textbf{F[Hz]} & \textbf{Subjects}                                                               & \begin{tabular}[c]{@{}c@{}}\textbf{Subjets with}\\\textbf{Multiple Sessions}\end{tabular}                                                               & \textbf{Recording}      \\ \hline
\hline
\begin{tabular}[c]{@{}c@{}}In-house\\(only SR)\end{tabular} & 12    & 500 & 81,974                                                                & 26,007 & On-TP  \\ \hline
PTB      & 12   & 1K & \begin{tabular}[c]{@{}c@{}}290 \\ (52 healthy)\end{tabular}            & \begin{tabular}[c]{@{}c@{}}113 \\ (14 healthy)\end{tabular}    & On-TP  \\ \hline
ECG-ID   & 1 & 500    & 90                                                                     & 89     & On-TP  \\ \hline
CYBHi    & 1 & 1K    & \begin{tabular}[c]{@{}c@{}}63 \end{tabular} & 63                                                               & Off-TP \\ \hline
\end{tabular}
\end{adjustbox}
\label{table:databases}
\end{table}

\begin{table*}[tb]
\centering
\caption{Experimental protocol considered for the training of the Siamese verification system, and the final single- and multi-session evaluation.}
\scalebox{1.00}{\renewcommand{\arraystretch}{1.2}%
\begin{tabular}{cccccccc}
\begin{tabular}[c]{@{}c@{}}\textbf{Verification}\\\textbf{Phases}\end{tabular} & \textbf{Database}       & \begin{tabular}[c]{@{}c@{}}\textbf{Enrolment}\\\textbf{Segment}\end{tabular}                 & \begin{tabular}[c]{@{}c@{}}\textbf{Enrolment}\\\textbf{Session}\end{tabular} & \begin{tabular}[c]{@{}c@{}}\textbf{Probe}\\\textbf{Segment}\end{tabular}           & \begin{tabular}[c]{@{}c@{}}\textbf{Probe}\\\textbf{Sessions}\end{tabular} & \begin{tabular}[c]{@{}c@{}}\textbf{Genuine}\\\textbf{Comparisons}\end{tabular}          & \begin{tabular}[c]{@{}c@{}}\textbf{Impostor}\\\textbf{Comparisons}\end{tabular} \\ \hline \hline
Training                   &    \begin{tabular}[c]{@{}c@{}}In-house\\(multi-session group)\end{tabular}                                      & Template                                                                    & $\mathbf{1^{st}}$                                                         &     \begin{tabular}[c]{@{}c@{}}Single\\Segment\end{tabular}                                                                                              & $\mathbf{2^{nd}}$, $\mathbf{3^{rd}}$, $\mathbf{4^{th}}$                                                   & up to \textbf{3} & \begin{tabular}[c]{@{}c@{}}\textbf{15}, from different\\ random subjects\end{tabular} \\ \hline
\begin{tabular}[c]{@{}c@{}}Single-session \\ Evaluation\end{tabular} & \begin{tabular}[c]{@{}c@{}}In-house (multi-session),\\PTB, ECG-ID, CYBHi\end{tabular}                                     & \begin{tabular}[c]{@{}c@{}}Summary\\Segment\end{tabular} & $\mathbf{2^{nd}}$                                                         & \begin{tabular}[c]{@{}c@{}}Summary\\Segment\end{tabular} & $\mathbf{2^{nd}}$                                                             & \textbf{3}                                                                      & \begin{tabular}[c]{@{}c@{}}\textbf{15}, from different\\ random subjects\end{tabular} \\ \hline
\begin{tabular}[c]{@{}c@{}}Multi-session\\ Evaluation\end{tabular}   & \begin{tabular}[c]{@{}c@{}}In-house (multi-session),\\PTB, ECG-ID, CYBHi\end{tabular}        & Template                                                                    & $\mathbf{1^{st}}$                                                         & Template & $\mathbf{2^{nd}}$                                                             & \textbf{1}                                                                      & \begin{tabular}[c]{@{}c@{}}\textbf{5}, from different\\random subjects\end{tabular}  \\ \hline
\end{tabular}
}
\label{table:experimental_protocol_verification}
\end{table*} 

\subsubsection{\textbf{In-house Database}}
The first database is an in-house collection of 295,649 12-lead ECG signals recorded with the Philips 12-lead machine (\url{https://philips.to/36CPabZ}) from a cohort of 122,622 subjects at La Princesa University Hospital (Madrid, Spain), with approval by the Clinical Ethics Committee. 
We exclude from the study all the ECGs not recorded during ``sinus-rhythm''. Then, we divide the 138,706 remaining ECGs in two groups: \textit{i)} those belonging to subjects with only a single ECG recorded (single-session acquisition, 55,967 ECGs from 55,967 subjects, age $54.88 \pm 20.15$, $52.92\%$ women), and \textit{ii)} those belonging to subjects with two or more ECGs recorded (multi-session acquisition, 82,739 ECGs from 26,007 subjects, age $64.44 \pm 17.90$, $49.74\%$ women, average distance of $559.65 \pm 626.68$ days between the first two ECGs).

\subsubsection{\textbf{PTB}}
This database is collected from Physikalisch-Technische Bundesanstalt (Germany) with a non-commercial prototype recorder~\cite{bousseljot_nutzung_2009}. The database contains between one to five ECG signals from 290 subjects (209 men, mean age $57.2$). In our experiments, we consider signals related to both 12 standard leads and single Lead I. We identify two not-disjoint sets of subjects: \textit{i)} those with multiple recordings (113 subjects), and \textit{ii)} those considered healthy (52 subjects). In the literature, both sets have been considered for experiments. PTB is one of the biggest public 12-lead ECG databases suitable for multi-session acquisition analysis. 

\subsubsection{\textbf{ECG-ID}}
This single-lead (Lead I) database is collected with limp clamp electrodes that imitate the scenario of user interaction with practical identification systems~\cite{lugovaya2005}. ECG signals are recorded from 90 volunteers (44 men, aged from 13 to 75). The number of ECGs for each subject varies from 2, collected during one day, to 20, collected over 6 months, except for a subject with a single ECG signal that we exclude from the analysis. In multi-session analysis, we consider the first two ECGs for each subject to compare our results with other studies.

\subsubsection{\textbf{CYBHi}}
This is an example of ``off-the-person'' database, where ECG signals are recorded with dry Ag/AgCl electrodes~\cite{da_silva_check_2014}. ECG signals present virtually the same morphology of Lead I derivation of a standard 12-lead medical ECG. The database contains two datasets: \textit{i)} short-term, with multiple ECG signals recorded in a five minutes-period from 65 healthy participants (49 men, age $31.1 \pm 9.46$) subject to different external stimuli, \ie low and high arousal videos, and \textit{ii)} long-term, with two ECG signals recorded with three months-distance from 63 healthy participants (14 men, age $20.68 \pm 2.83$). We consider ECG signals from the long-term dataset to address the most challenging scenario. CYBHi presents numerous challenges compared to other databases, due to the lower signal-to-noise ratio of its ECG signals.

\section{Experimental Work: Verification Task}\label{sec:verification}

\subsection{Experimental Protocol}
We consider our large-scale in-house database, divided in two groups as specified in Sec.~\ref{sec:databases}. Data from the first group (single-session acquisition) are used to train the Autoencoder. Data from the second group (multi-session acquisition) are used to perform experiments in ECG biometric recognition, as they provide multiple ECG signals for each subject. In this study we perform cross-dataset evaluation, as we train ECGXtractor with the in-house database and evaluate it with several additional databases.

At first, we train our Autoencoder with the single segments extracted from the ECG signals contained in the first group, composed of 55,967 ECGs that we divide into training and validation sets, with 80:20 ratio. Up to three single segments of each ECG signal have been considered, having in total 133,575 single segments in the training set and 33,357 single segments in the validation set. We use mean squared error as loss function, with Adam optimizer and initial learning rate of 0.001. At each epoch we evaluate the loss function on the validation set. We halve the learning rate if the function does not decrease for two consecutive epochs, and we stop the training if the function does not decrease for six consecutive epochs.

Then, we focus on the training and evaluation of our 1L- and 12L-Siamese CNNs, for which we consider the same settings specified for the training of the Autoencoder, with cross-entropy as loss function. We employ the ECG signals contained in the second group, composed of 26,007 subjects that we divide into training, validation, and test sets, according to 70:10:20 ratio. The same test subjects are considered for both single- and multi-session acquisition scenarios. 

Given that subjects are randomly selected to generate impostor pairs, we evaluate our two Siamese CNNs ten times for each scenario and database considered. The results reported in this study are the averages of the values obtained in the ten executions of each specific scenario and database. Details of the experimental protocol considered for the training of the Siamese verification system, and the final single- and multi-session evaluation are provided in Table~\ref{table:experimental_protocol_verification} and discussed in the following. We note that in single-session evaluation, enrolment and probe segments are obtained from the second session of each subject, for a better comparison with multi-session scenario. Also, in multi-session evaluation we consider only one genuine comparison for each subject, as many subjects only have two ECG signals, and both enrolment and probe templates are obtained from whole signals. We provide next more details regarding the training of the Siamese verification system and the final single- and multi-session evaluation:


\begin{itemize}

 \item \textbf{\textit{Training:}} In total, we consider 54,351 genuine and 271,755 impostor pairs to train our two Siamese CNNs (\ie 1-Lead and 12-Lead scenarios). We use the same pairs for both Siamese CNNs, with the only difference consisting in the number of leads.

 \item \textbf{\textit{Single-session Evaluation:}} For each subject, three genuine pairs and fifteen impostor pairs are generated by randomly matching summary segments of the same and different subjects. We verify that each generated pair contains different summary segments, and that the same pair is not considered multiple times. To evaluate our Siamese CNNs with the in-house database, we consider the test subjects not previously used for training and validation. With the in-house and PTB databases, we can evaluate both 1L- and 12L-Siamese CNNs. To obtain comparable results, the same genuine and impostor pairs are considered for the two Siamese CNNs. Additionally, we also take into account the subset of healthy subjects in the PTB database. Given that 38 of the 52 healthy subjects are provided with a single ECG signal, we consider the subset only for single-session scenarios.
 
 \item \textbf{\textit{Multi-session Evaluation:}} For each subject we select two distinct ECG signals and generate an enrolment template from the first session, and a probe template from the second one. Then, we create genuine pairs by matching the two templates of each subject, and impostor pairs by matching the enrolment template of each subject with five probe templates of different subjects. The same subjects considered for the evaluation of single-session verification are used to evaluate our Siamese CNNs in multi-session. Also, the same comparison pairs are used to evaluate 1L- and 12L-Siamese CNNs with in-house and PTB databases.

\end{itemize}

To sum up, we underline that in single-session scenario only experiments with summary segments can be performed, as we dispose of a single ECG signal for each subject. Nevertheless, in multi-session scenario we dispose of multiple ECG signals for each subject, and we observed that templates instead of summary segments provide better performances.

\subsection{Experiment 1: Single-Session Verification}

In Table~\ref{table:single-session_verification} we report the performance achieved in terms of EER, along with the number of genuine and impostor pairs tested. It is important to highlight that only the in-house database has been considered for training ECGXtractor. Therefore, in Table~\ref{table:single-session_verification} we can also analyze the generalization ability of ECGXtractor to other databases and sensors. We observe better EERs when we consider 12-lead instead of 1-lead ECG signals, \ie 1.28\% vs 3.27\% EER for the in-house database and 0.14\% vs 0.42\% EER for PTB. Also, the two Siamese CNNs are able to provide perfect 0\% EER for the ECG signals belonging to healthy subjects of PTB, and 1.52\% EER with ECG-ID database, better compared to the results of~\cite{chu_ecg_2019} and~\cite{ingale_ecg_2020} presented in Table~\ref{table:related_works}. We observe that our two Siamese CNNs, trained with the in-house database and evaluated with other databases, provide a remarkable generalization ability in cross-dataset evaluation.

Our 1L-Siamese CNN shows performance degradation when evaluated with comparison pairs obtained from CYBHi database (6.98\%). The availability of only a single lead, the high variability between heartbeats recorded during the same session, and the low signal-to-noise ratio, peculiar of this database, may be the causes of the performance degradation. We observe that~\cite{ingale_ecg_2020} achieves values ranging from 2.3\% to 9\% EER for CYBHi. However, that study considers a larger set of subjects, some of them provided with ECG signals recorded under different conditions, and the protocol adopted to generate comparison pairs is not clearly specified. In general, worse values of EER for CYBHi database are common in the literature, compared to EERs achievable with other databases. For instance, high EERs are observable in~\cite{da_silva_luz_learning_2018} (up to 26.38\%) and~\cite{pinto_end--end_2019} (15.37\%).

\begin{table}[tb]
\centering
\caption{Description of the different evaluation sets and performances achieved in the scenario of \textbf{single-session ECG biometric verification}. EER = Equal Error Rate. }
\begin{adjustbox}{width=0.455\textwidth}\renewcommand{\arraystretch}{1.2}%
\begin{tabular}{ccccc}
\textbf{Database}      & \begin{tabular}[c]{@{}c@{}}\textbf{Genuine}\\\textbf{Comparisons}\end{tabular} & \begin{tabular}[c]{@{}c@{}}\textbf{Impostor}\\\textbf{Comparisons}\end{tabular} & \textbf{Leads} & \textbf{EER {[}\%{]}} \\ \hline\hline
\multirow{2}{*}{In-house}       & \multirow{2}{*}{15,534}         & \multirow{2}{*}{77,670}          & 12    & 1.28         \\ \cline{4-5}
              &               &                & 1     & 3.27         \\ \hline
\multirow{2}{*}{PTB}           & \multirow{2}{*}{339}           & \multirow{2}{*}{1,695}           & 12    & 0.14 \\ \cline{4-5}
              &               &                & 1     & 0.42         \\ \hline
\multirow{2}{*}{\begin{tabular}[c]{@{}c@{}}PTB\\(healthy)\end{tabular}} & \multirow{2}{*}{156}           & \multirow{2}{*}{780}            & 12    & 0.00         \\ \cline{4-5}
              &               &                & 1     & 0.00         \\ \hline
ECG-ID        & 267           & 1,335           & 1     & 1.52         \\ \hline
CYBHi         & 189           & 945            & 1     & 6.98        \\\hline
\end{tabular}
\end{adjustbox}
\label{table:single-session_verification}
\end{table}

\subsection{Experiment 2: Multi-Session Verification}

In Table~\ref{table:multi-session_verification} we report the performance achieved in terms of EER, along with the number of genuine and impostor pairs tested, and the average distance in days between the ECG signals considered for each subject. Again, as described in the single-session experiment, only the in-house database has been considered for training ECGXtractor. Therefore, in Table~\ref{table:multi-session_verification} we can also analyze the generalization ability of ECGXtractor to other databases and sensors.

Analysing the results of this scenario, we observe that they are generally worse than those achieved in single-session verification, for both single- and multi-lead ECG signals. In particular, the largest performance degradation affects the single-lead scenario for PTB database, with a worsening of EER from 0.42\% to 5.12\%.

The exception to this trend is represented by ECG-ID, that provides a very low EER of 0.15\% for multi-session verification. We highlight that the ECG signals considered for each subject are recorded during the same day. Moreover, the possibility to average the entire amount of heartbeats to generate templates in multi-session verification may improve the performance previously achieved in single-session verification with summary segments. However, some variations are expected between ECG signals recorded during the same day but in different sessions, because of heart rate variability~\cite{islam_selection_2017} and potential alterations introduced by the operator between different measurements. 

It is not possible to make a fully comparative analysis with other studies in Table~\ref{table:related_works}, due to different experimental protocols and missing details. Several studies do not perform multi-session analysis, or consider databases that provide a single ECG for many subjects. This is one of the main motivations of the present study, \ie to establish a benchmark evaluation that is easily reproducible by the research community.

\begin{table}[tb]
\centering
\caption{Description of the different evaluation sets and performances achieved in the scenario of \textbf{multi-session ECG biometric verification}. EER = Equal Error Rate, FT = Fine Tuning.}
\begin{adjustbox}{width=0.499\textwidth}\renewcommand{\arraystretch}{1.4}%
\begin{tabular}{cccccc}

\textbf{Database} & \begin{tabular}[c]{@{}c@{}}\textbf{Genuine}\\\textbf{Comp.}\end{tabular} & \begin{tabular}[c]{@{}c@{}}\textbf{Impostor}\\\textbf{Comp.}\end{tabular} & \begin{tabular}[c]{@{}c@{}}\textbf{Distance}\\\textbf{[Days]}\end{tabular} & \textbf{Leads} & \textbf{EER {[}\%{]}}  \\ \hline\hline
\multirow{2}{*}{In-house}  & \multirow{2}{*}{5,162}          & \multirow{2}{*}{25,801}        & \multirow{2}{*}{566 $\pm$ 631}  & 12    & 1.97                 \\ \cline{5-6}
         &               &            &    & 1     & 5.55                \\ \hline
\multirow{2}{*}{PTB}      & \multirow{2}{*}{113}           & \multirow{2}{*}{565}        & \multirow{2}{*}{23 $\pm$ 146}       & 12    & 2.06               \\ \cline{5-6}
         &               &           &     & 1     & 5.12                 \\ \hline
ECG-ID   & 89            & 445          & same day  & 1     & 0.15         			          \\ \hline
CYBHi    & \multirow{2}{*}{63}            & \multirow{2}{*}{315}        & \multirow{2}{*}{105 $\pm$ 5}    & 1     & 7.97                	  \\ \cline{1-1} \cline{5-6}
CYBHi (FT)   &               &         &       & 1     & 5.44               	  \\ \hline
\end{tabular}
\end{adjustbox}
\label{table:multi-session_verification}
\end{table}

\subsection{Experiment 3: Fine tuning}

To conclude the analysis of multi-session verification, we investigate the possibility to fine tune ECGXtractor with data from specific databases. In this study, we fine tune our 1L-Siamese CNN with comparison pairs obtained from the ECG signals contained in the short-term dataset of CYBHi, not previously considered for evaluation. We exclude 10 of the 65 subjects available in short-term dataset, as they are also included in the long-term dataset used for evaluation. We divide the remaining subjects into training and validation sets with 80:20 ratio, and extract single segments from their ECG signals. With the usual 1:5 genuine-impostor ratio, we generate 12,980 comparison pairs for training and 3,058 pairs for validation. To increase variability during fine tuning, each pair is made of two single segments. We evaluate the fine-tuned 1L-Siamese CNN with the same comparison pairs previously used for multi-session verification with CYBHi. Table~\ref{table:multi-session_verification} shows the results achieved over the evaluation dataset, \ie CYBHi (FT). We obtain an improvement of EER from 7.97\% to 5.44\%. Similar results would be expected when doing similar procedure with other databases.

\section{Experimental Work: Identification Task}\label{sec:identification}

\subsection{Experimental Protocol}

For the scenario of identification, we only perform experiments with PTB and ECG-ID databases, using summary segments generated from blocks of ten consecutive single segments. Given the different number of summary segments available between subjects, we perform random re-sampling to train and evaluate each system with the same number of summary segments for each subject. Details of the experimental protocol considered for the training and evaluation of the identification system in different scenarios are provided in Table~\ref{table:experimental_protocol_identification}.
 
\begin{itemize}
	\item \textbf{\textit{Training:}} For each experiment, we consider our Autoencoder for feature extraction and the singular Siamese component of 1L- or 12L-Siamese CNNs, without further training them. The only layer trained is the final fully connected layer, included at the end of the Siamese component and specific to each experiment (\ie depending on the number of users to identify). We use categorical cross-entropy as loss function and maintain the other settings specified for the previous training.
	\item \textbf{\textit{Single-session:}} We perform two different experiments involving subsets of PTB database: \textit{i)} the 113 subjects provided with multiple ECG signals, to compare performance between single- and multi-session, and \textit{ii)} the 52 healthy subjects, to compare the achieved performance with other studies in the literature. In both experiments, we select a single ECG signal for each subject, generate all the possible summary segments, and split them in the training, validation, and test sets with 70:10:20 ratio. 
	\item \textbf{\textit{Multi-session:}} For multi-session, we consider the subset of 113 subjects in PTB provided with at least two ECG signals. For each subject, we generate all the possible summary segments from all but the last of their ECG signals. We split the summary segments of each subject in training and validation sets with 80:20 ratio. Then, we generate all the possible summary segments from the last ECG signal of each subject and use them for evaluation.
We also consider the ECG-ID database. To increase the amount of available data, we mix together the summary segments generated from the first two ECG signals of each subject. We call this scenario ``mixed-session'', to be more precise. We split the summary segments of each subject in the training, validation, and test sets with 70:10:20 ratio. 
\end{itemize}

\begin{table}[tb]
\centering
\caption{Experimental protocol considered for the training and evaluation of the identification system in the scenario of single-, multi-, and mixed-session.}
\scalebox{1.00}{\renewcommand{\arraystretch}{1.2}%
\begin{tabular}{cccccc}
\textbf{Scenario}              & \textbf{DB} & \textbf{\begin{tabular}[c]{@{}c@{}}Tr./Eval. \\ Segment\end{tabular}} & \textbf{\begin{tabular}[c]{@{}c@{}}Training\\ Session\end{tabular}} & \textbf{\begin{tabular}[c]{@{}c@{}}Evaluation\\ Segment\end{tabular}} & \textbf{\begin{tabular}[c]{@{}c@{}}Tr. Sample\\ per Subject\end{tabular}} \\ \hline \hline
\begin{tabular}[c]{@{}c@{}}Single-\\session\end{tabular}                 & \begin{tabular}[c]{@{}c@{}}PTB\\(healthy)\end{tabular}               & \begin{tabular}[c]{@{}c@{}}Summary\\ Segment\end{tabular}             & $1^{st}$           & $1^{st}$                                                                     & \begin{tabular}[c]{@{}c@{}}18\\(14)\end{tabular} \\ \hline
\begin{tabular}[c]{@{}c@{}}Multi-\\session\end{tabular} & PTB               & \begin{tabular}[c]{@{}c@{}}Summary\\ Segment\end{tabular}             & \begin{tabular}[c]{@{}c@{}}All but\\the last\end{tabular}             & Last                                                              &     68                                                                           \\ \hline
\begin{tabular}[c]{@{}c@{}}Mixed-\\session\end{tabular}                               & \begin{tabular}[c]{@{}c@{}}ECG-\\ID\end{tabular}           & \begin{tabular}[c]{@{}c@{}}Summary\\ Segment\end{tabular}             & $1^{st}$, $2^{nd}$             & $1^{st}$, $2^{nd}$ & 5      \\ \hline
\end{tabular}
}
\label{table:experimental_protocol_identification}
\end{table}

\begin{table}[tb]
\centering
\caption{Description of the different evaluation sets and performance achieved in the scenario of \textbf{single-, multi-, and mixed-session ECG biometric identification}.}
\begin{adjustbox}{width=0.455\textwidth}\renewcommand{\arraystretch}{1.2}%
\begin{tabular}{cccccccc}
\textbf{Database}      & \textbf{Leads} & \textbf{Session} & \textbf{Subjects}   & \textbf{Accuracy {[}\%{]}} \\ \hline\hline
PTB           & 12    & Single  & 113     & 100               \\ \hline
PTB (healthy) & 12    & Single  & 52      & 100               \\ \hline
PTB           & 12    & Multi   & 113     & 96.46             \\ \hline   
ECG-ID        & 1     & Mixed   & 89      & 97.75             \\ \hline             
\end{tabular}
\end{adjustbox}
\label{table:identification}
\end{table}

\subsection{Experiment 1: Single-Session Identification}
In Table~\ref{table:identification} we report the performance achieved in terms of accuracy. Compared with the verification scenario, each subject of PTB is correctly identified, also when considering the subset of 113 subjects. This result is consistent with those achieved in the literature~\cite{alduwaile_using_2021, chu_ecg_2019} and reported in Table~\ref{table:related_works}.

\subsection{Experiment 2: Multi-Session Identification}
In Table~\ref{table:identification} we also report the accuracy achieved for multi- (and mixed-) session. As in the case of verification, the performance decreases between single- and multi-session, with accuracy that goes from 100\% to 96.46\% for PTB. Higher accuracies in multi-session scenario are achieved in recent studies specifically designed for identification: 99.66\% for PTB~\cite{srivastva_plexnet_2021}, and 100\% for ECG-ID~\cite{ihsanto_fast_2020}. We remind that identification is not the main focus of this study, and that these results are achieved with an architecture of our ECGXtractor trained for verification with the in-house database, while other databases are used for evaluation. No fine tuning is considered here, and only minor adaptations of our system have been made, proving the potential of ECGXtractor to extract discriminative ECG-based features for different tasks and scenarios.

\section{Conclusion}\label{sec:conclusion}
In this article we have investigated ECG biometric recognition through the proposal of ECGXtractor, a novel DL method, exhaustively evaluated across multiple scenarios and precise experimental settings. We release in GitHub the trained weights of our DL system, along with the material used to carry out our experiments\footnote[1]{\url{https://github.com/BiDAlab/ECGXtractor}}. This study aims to overcome the major drawback of ECG biometric recognition, \ie the lack of standard experimental protocols, by making available a general public benchmark evaluation so that everyone can replicate it and compare their results with us. The proposed ECGXtractor method is robust in cross-dataset evaluation, with the following best EERs achieved in multi-session verification: 1.97\% for the in-house database, 2.06\% for PTB, 0.15\% for ECG-ID, and 5.44\% for CYBHi. Moreover, ECGXtractor performs perfectly when the subset of healthy subjects from PTB database is considered, in both single-session verification and identification.

In addition, the evaluation conducted with the ``off-the-person'' CYBHi database may present interesting implications. ECG signals like those of CYBHi are the easiest to record in widespread applications, and studies involving ``off-the-person'' databases may favour the diffusion of ECG biometric recognition technologies. The major drawbacks of these signals are the high level of noise and imprecision, that generally lower recognition performance compared to traditional ECGs. In this sense, a strategy analysed in this study and requiring further investigation is the fine-tuning of ECGXtractor with specific ECG signals presenting the characteristics of the signals of interest. For CYBHi, we were able to decrease the EER in multi-session verification from 7.97\% to 5.44\%.

Another future work consists in the protection of sensitive information that may be contained in the different types of ECG segments considered in this study~\cite{melzi_analyzing_2021}. As we observed, the Autoencoder of ECGXtractor was not trained specifically for recognition tasks. Hence, the features extracted from ECG segments are generic, and we expect that they can be successfully exploited in other applications, revealing sensitive information such as age, sex, or medical pathologies~\cite{delgado_2022}. The risk assessment related to this aspect and eventual countermeasures shall be further investigated.

\section*{Acknowledgments}
Many thanks to Instituto de Investigacion Sanitaria del Hospital Universitario de La Princesa, Madrid, Spain, and in particular to Guillermo J. Ortega, Luis Jesus Jimenez-Borreguero, Alberto Cecconi and Ancor Sanz-Garcia for granting us the access to the in-house database used in this study. 

This project has received funding from the European Union’s Horizon 2020 research and innovation programme
under the Marie Sklodowska-Curie grant agreements No 860813 - TRESPASS-ETN and 860315 - PRIMA-ITN. 

This work has been supported by projects: IDEA-FAST (IMI2-2018-15-853981), and INTER-ACTION (PID2021-126521OB-I00 MICINN/FEDER).

{
\bibliographystyle{IEEEtran}
\bibliography{egbib2}
}

\end{document}